\documentclass[a4paper]{article}

\usepackage{INTERSPEECH2022}
\usepackage{multirow}
\usepackage{color}
\usepackage{subfigure}
\usepackage{makecell}
\usepackage{CJK}
\usepackage{arydshln}
\usepackage{algorithm}
\usepackage{algorithmic}

\title{On the Effectiveness of Pinyin-Character Dual-Decoding for End-to-End Mandarin Chinese ASR}
\name{Zhao Yang$^{1,3}$, Dianwen Ng$^{2,3}$, Xiao Fu$^{1}$, Liping Han$^{1}$, Wei Xi$^1$, Rui Wang$^1$, Rui Jiang$^1$,\\ Jizhong Zhao$^1$}
\address{
  $^1$Faculty of Electronic and Information Engineering, Xi’an Jiaotong University, China \\
  $^2$Alibaba Group, Singapore, Singapore \\
  $^3$School of Computer Science and Engineering, Nanyang Technological University, Singapore}
\email{\{zhaoyang9425, ruijiang.jerry, xiaofu9804, hlpofwork\}@gmail.com, dianwen.ng@alibaba-inc.com, \{xiwei, zjz\}@xjtu.edu.cn, wangrui1203@stu.xjtu.edu.cn}

\begin{document}

\maketitle
\begin{abstract}
 End-to-end automatic speech recognition (ASR) has achieved promising results. However, most existing end-to-end ASR methods neglect the use of specific language characteristics. For Mandarin Chinese ASR tasks, there exist mutual promotion relationship between Pinyin and Character where Chinese characters can be romanized by Pinyin. Based on the above intuition, we first investigate types of end-to-end encoder-decoder based models in the single-input dual-output (SIDO) multi-task framework, after which a novel asynchronous decoding with fuzzy Pinyin sampling method is proposed according to the one-to-one correspondence characteristics between Pinyin and Character. Furthermore, we proposed a two-stage training strategy to make training more stable and converge faster. The results on the test sets of AISHELL-1 dataset show that the proposed enhanced dual-decoder model without a language model is improved by a big margin compared to strong baseline models.
\end{abstract}

\section{Introduction}

End-to-end automatic speech recognition (ASR) \cite{graves2012sequence,chiu2018state,bahdanau2016end,nakatani2019improving,kim2017joint} has recently drawn an extensive research interest for its sleek and simplistic design compared to the traditional ASR system with multiple independent components. Besides, neural network models based on connectionist temporal classification (CTC) \cite{hannun2017sequence,tian2019self}, networks attention \cite{bahdanau2016end,chan2016listen,shan2019component} and Transformer \cite{winata2020lightweight,dong2018speech,li2019speechtransformer} have received numerous successes in the ASR task for assorted languages. Nonetheless, these ASR models often overlook the presence of the intrinsic association between the character and spelling, especially in the case of the Chinese language, where we could potentially utilize to improve the performance of our transcription tasks. To elaborate, the Chinese language is notoriously difficult to learn. The ideographic nature of the language creates a unique symbol to every word or morpheme that does not directly express the pronunciation \cite{chan2016online}, making the identification of the Chinese Character from utterance challenging. However, the Chinese language contains a complementary writing system called Pinyin \cite{zhang2020pinyin} that characterize the text Characters by the romanized spelling. It uses the phonetic system with the Latin alphabet to represent each word in a sentence that provides the intermediate linkage from the vocal pronunciation to the text Character. This establishes the connection of a one-to-one correspondence between the graphemes and phonemes. As such, integrating the information of Pinyin in Mandarin Chinese ASR is likely to foster efficacious support in recognizing the actual text of a given speech. Furthermore, many studies \cite{Chung2003EffectsOP, Du2011PinyinAC} have also shown on human learners that infusing Pinyin in their learning significantly improves English-Chinese bilingual learners to instil new Chinese vocabulary at the beginner level. Therefore, this strongly motivates us to extend the recent ASR technology for Mandarin Chinese (ideographic language) to include the awareness of Pinyin when executing the recognition task. This involves the design of an end-to-end joint Pinyin-Character model. Meanwhile, displaying Pinyin transcripts alongside the character transcripts is more friendly to users.

As far as we know, related works using Pinyin to perform the Mandarin Chinese ASR task \cite{chan2016online, zou2018comparable, zhou2018syllable, wang2021cascade} appears to be infrequently investigated. Additionally, existing approaches can be categorized into the two: \textit{a) Cascading} \cite{wang2021cascade, zhang2019investigation, fu2020research}. A model is first used to transform acoustic features into a sequence of Pinyin syllables. Then, the sequence of syllables is converted to the Character through a syllable-to-character converter. \textit{b) End-to-end}. \cite{chan2016online} proposed a joint Character-Pinyin model, in which the architecture for Pinyin generating and Character generating tasks are sharing the same encoder-decoder structure.

Character as one of the most common modeling units is chosen for Chinese ASR task. In this paper, we first investigate several attention-based encoder-decoder (AED) models which conform to single speech input and bilateral dual text output. These models leverage the readily available Pinyin information while concurrently emitting sequence of Characters as the final output.
Then, according to the one-to-one correspondence between Pinyin and Character targets, we design a new end-to-end dual-decoder asynchronous decoding method by glimpsing future Pinyin information. For each time step of Character decoding, the Character decoder decodes the current output through Characters from the past and Pinyin syllables from the future. Moreover, to reduce the differences in data distribution between the training and inference when fusing future Pinyin syllables with characters, generated Pinyin syllables are either from the ground truth or the generated syllables from confusion set with a certain probability.
Last but not least, we also design an evaluation metric to more accurately measure mutual promotion ability by comparing the alignment degree between dual outputs.

\begin{figure*}[htbp]
	\subfigure[]{
		\begin{minipage}[b]{0.18\linewidth}
			\centering
			\centerline{\includegraphics[width=2.8cm]{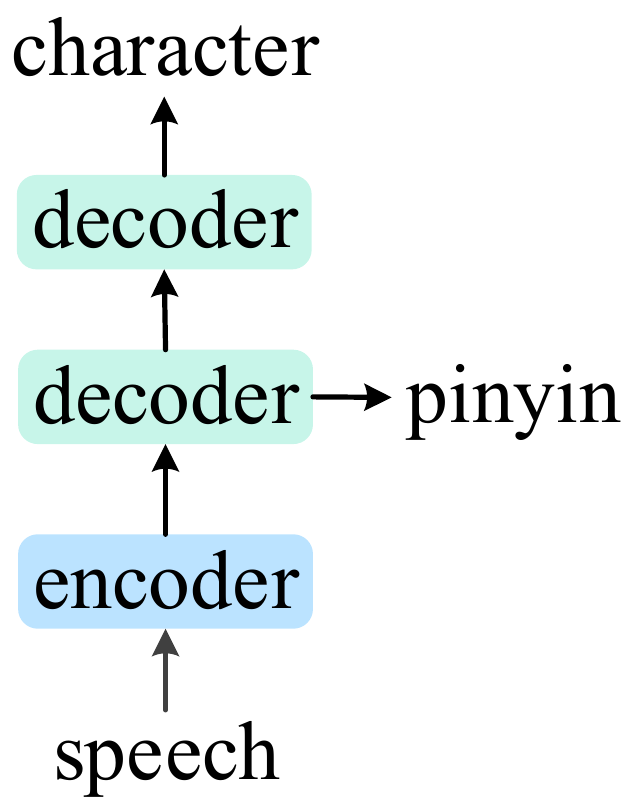}}
			\label{cascademodel}
		\end{minipage}
	}
	\hfill
	\subfigure[]{
		\begin{minipage}[b]{0.18\linewidth}
			\centering
			\centerline{\includegraphics[width=2.5cm]{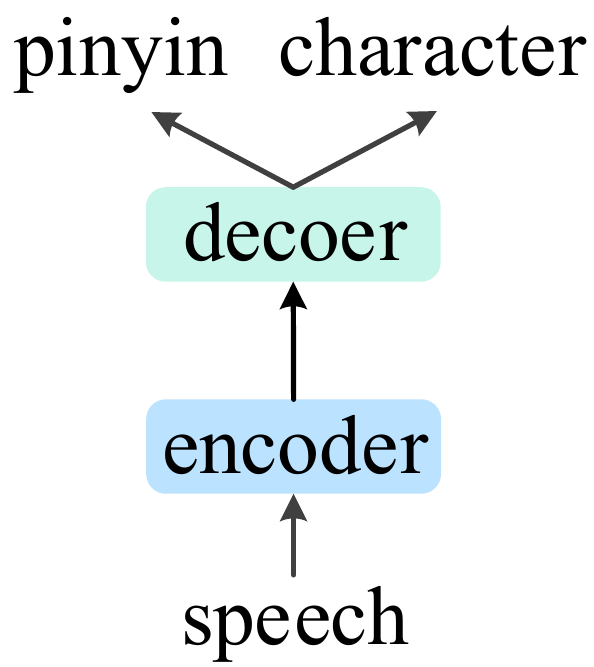}}
			\label{singlecharacter}
		\end{minipage}
	}
	\hfill
	\subfigure[]{
		\begin{minipage}[b]{0.19\linewidth}
			\centering
			\centerline{\includegraphics[width=3.5cm]{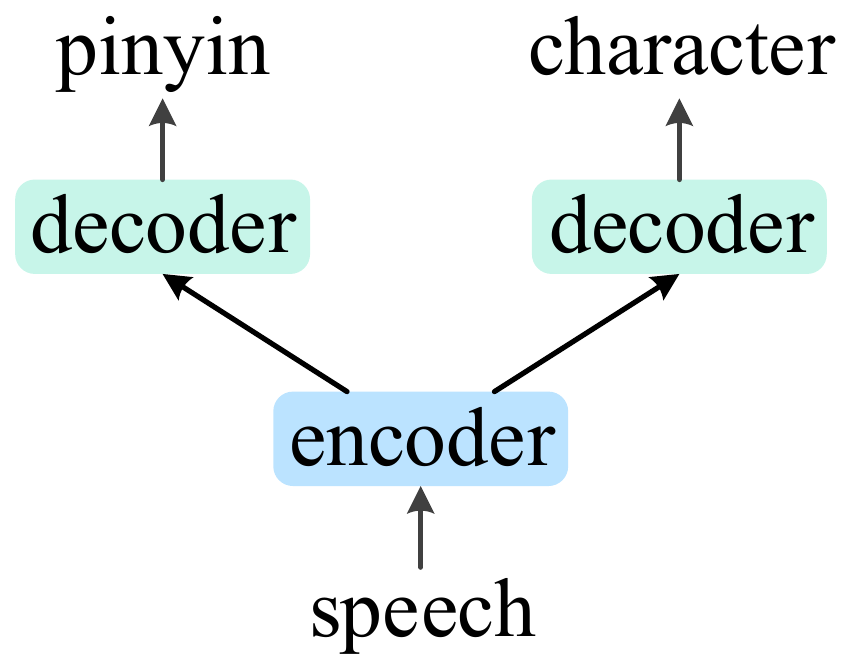}}
			\label{parallelmodel}
		\end{minipage}
	}
	\hfill
	\subfigure[]{
		\begin{minipage}[b]{0.18\linewidth}
			\centering
			\centerline{\includegraphics[width=2.8cm]{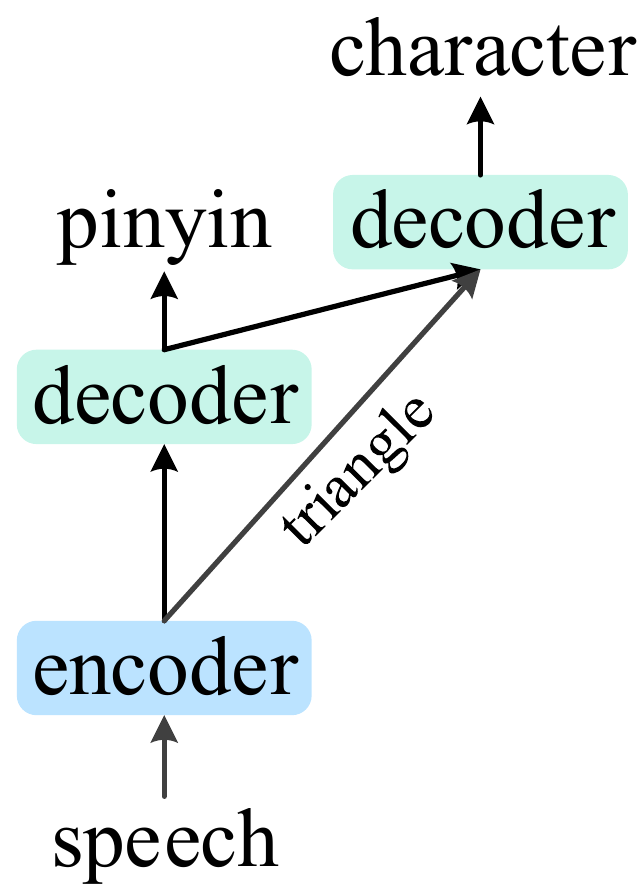}}
			\label{trianglemodel}
		\end{minipage}
	}
	\hfill
	\subfigure[]{
		\begin{minipage}[b]{0.19\linewidth}
			\centering
			\centerline{\includegraphics[width=3.5cm]{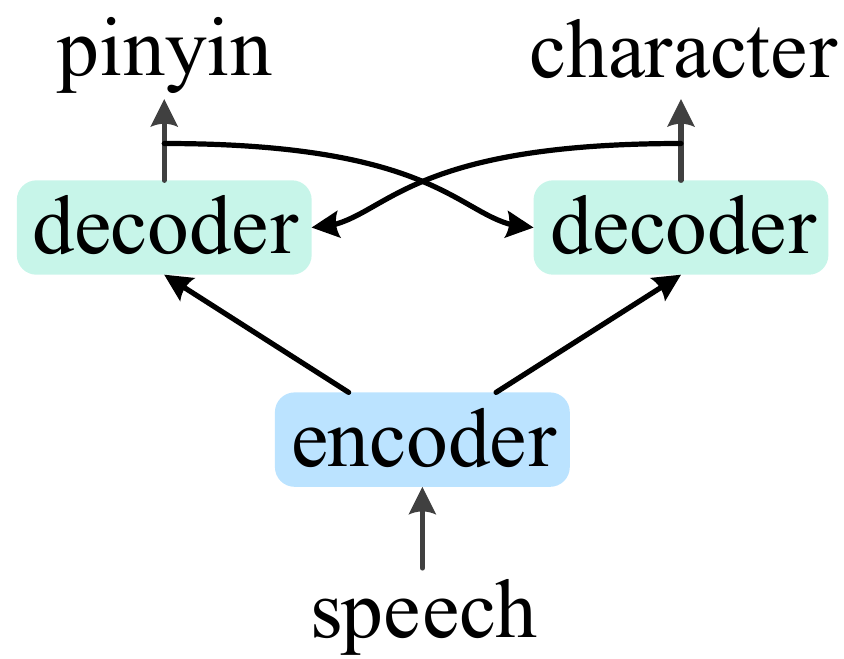}}
			\label{interactivemodel}
		\end{minipage}
	}
	\caption{Diagram of the investigating Pinyin-Character SIDO ASR models. (a) Hierarchical-stage. (b) Multi-task. (c) Multi-decoder. (d) Triangle. (e) Interactive multi-decoder.}
	\label{relatedmodels}
	\vspace{-0.5cm}
\end{figure*}

\section{A Study of Pinyin-Chinese ASR}

This section introduces variants of Mandarin Chinese ASR architecture that appear moderately in the literature of other works where we could use it to construct Pinyin awareness in our model. Here, we refer to some common approaches \cite{liu2020synchronous,le2020dual, anastasopoulos-chiang-2018-tied, chan2016online} with a designed computation for feature awareness to achieve the study objective. Essentially, we investigate several single-input dual-output (SIDO) frameworks to teach the awareness from one of the task outputs that guides the model to predict Pinyin from the given utterance. A succinct description for each architecture is offered in the following paragraphs. In addition, a diagram visualization of the list of model structures is presented in Figure \ref{relatedmodels}. 

\noindent\textbf{Hierarchical stage modelling} \cite{sogaard-goldberg-2016-deep}--We build a hierarchical model where the higher-level task has a partial dependency on the lower-level task. This architecture assumes a graphical dependency in the two defined tasks. In particular, since Pinyin and Character are connected by a one-to-one correspondence between the phonemes and graphemes where phonemes should naturally come before graphemes, we can construe the lower-level task as the recognition of Pinyin whilst the higher-level focuses on our main ASR task. As a result, the model first gains proficiency in learning the phonetic feature and share it in the latter task to boost its performance. The model will update its learning at the different intermediate layers as intended by the proposed hierarchical level of the decoder. 

\noindent\textbf{Multi-task modelling} \cite{chan2016online}--As opposed to the previous hierarchical stage modelling, we assume the absence of the hierarchical relationship by setting a multi-supervision task at the decoder to predict Pinyin and Character. The model adds two separate MLP layers at the output layer of our decoder to extract features that are rich in Pinyin and Character oriented.

\noindent\textbf{Multi-decoder modelling}--This architecture assumes the independent relationship of decoding from Pinyin and Character. The model learns separately and no information should be distributed across the two when inferring the Pinyin and Character. This allow our decoders to be specialized in their designated task. However, we enforce our model to share the same encoder where it extracts features that are rich in Pinyin and Character at the encoding level.

\noindent\textbf{Triangle modelling}  \cite{anastasopoulos-chiang-2018-tied}--This is a hybrid of the formal multi-decoder and Hierarchical stage modelling. Similarly, we assume the rank-order hierarchical relationship but inferring Pinyin should be executed separately from Character. Here, the character decoder has two cross-attention modules, one for the input states of the speech encoder and another for the output states of the pinyin decoder. The two attention modules are hierarchically connected. 

\noindent\textbf{Interactive multi-decoder modelling}--Since the dual-decoders are highly associated, we can assume both to well-complement each other. Hence, we can allow the two decoders to attend information from one another via a layer-wise dual-attention mechanism \cite{liu2020synchronous,le2020dual}. The cross-decoder interaction occurs in the self-attention sub-layer. According to the direction of interaction, it includes unilateral interaction and bilateral interaction. Take unilateral interaction as an example, it is first calculated to obtain cross-decoder representation:
\begin{equation}
\textbf{H}_{\text{cross-decoder}} = Attention(\textbf{Q}_{1}, \textbf{K}_{2}, \textbf{V}_{2})
\end{equation}
where $\textbf{Q}_{1}$ is query of hidden representation from Character (or Pinyin) self-attention sub-layer. $\textbf{K}_{2}$, $\textbf{V}_{2}$ are key and value of hidden representation from Pinyin (or Character) self-attention sub-layer.

Then the output of self-attention sub-layer and that of cross-decoder module can be integrated by a linear fusion function to obtain the final representation.
\begin{equation}
\textbf{H}_{final} = Linear(Concat(\textbf{H}_{\text{cross-decoder}}, \textbf{H}))
\end{equation}
where $\textbf{H}$ is output from Character self-attention sub-layer.



\section{Methodology}
 
The joint Pinyin-Character dual-decoder model is trained in the multi-task framework. Next, we will introduce three effective ways to further enhance the capabilities of the SIDO model. In this section, we will first detail dual-decoder asynchronous decoding method with fuzzy Pinyin sampling. Then a two-stage training strategy is introduced.
 
\subsection{Asynchronous Training with Fuzzy Pinyin sampling}

Since there exists one-to-one correspondence between Pinyin and Character output for single speech and the (nearly) symmetric dual-decoder structure, it is natural to consider prioritizing Pinyin decoding and then feed Pinyin information to Character decoder so that providing strongly correlated future Pinyin supplementary for Character decoding at each time step. In order to balance decoding speed and model effectiveness, the decoding time difference between dual decoders is set to 1 in this work. Given a speech sequence $\textit{x}= (\textit{x}_1, ..., \textit{x}_T)$, a Character target $\textit{y}^{char}= (\textit{y}_{1}^{char}, ..., \textit{y}_{T}^{char})$ and corresponding Pinyin target $\textit{y}^{pinyin}= (\textit{y}_{1}^{pinyin}, ..., \textit{y}_{T}^{pinyin})$, the Character decoding process can be formulated as:

\begin{equation}
\textit{p}(\textit{y}^{char} | \textit{x}) = \prod_{i=1}^{T} \textit{p}(\textit{y}_{i}^{char} | \textit{y}_{< i}^{char}, \textit{y}_{< i + 1}^{pinyin}, \textit{x})
\end{equation}

Asynchronous training is general but effective method for dual-decoder models with interaction between two decoders. Taking interactive multi-decoder category as an example, the asynchronous training is achieved by adjust mask in cross-decoder module.



Furthermore, the Pinyin embedding fed to Character decoder is from the ground truth during training while at inference the Pinyin syllables are generated by Pinyin decoder on its own. The discrepancy gap leads to terrible errors accumulation between training and inference. We address this issue by sampling context Pinyin syllables not only from the ground truth sequence but also from the generated Pinyin candidates from a confusion set which is a set of similar Pinyin syllables. Specifically, $\textit{p}$-percent syllables in each Pinyin training sample are randomly chosen to be replaced with similar ones from fuzzy Pinyin set.

From another perspective, asynchronous decoding enhances Characters generation by future Pinyin auxiliary, but making Pinyin generation isolated. Fuzzy Pinyin sampling can be regarded as a supplementary enhancement to Pinyin generation.

\begin{table*}[th]
	\begin{center}
			\setlength{\tabcolsep}{4.5mm}{
			\begin{tabular}{| c | c | c | c | c | c |}
				\hline
				\multirow{2}*{\textbf{Category}} & \multirow{2}*{\textbf{Model}} & \multicolumn{2}{c |}{\textbf{CER(\%) $\downarrow$}} & \multicolumn{2}{c|}{\textbf{Alignment Degree(\%) $\uparrow$}} \\
				\cline{3-6}
				& & \textbf{Pinyin} & \textbf{Character} & \textbf{Pinyin Pred} & \textbf{Pinyin GT}  \\
				\hline
				\multirow{2}*{Baseline} & PinyinOnly  & 3.13 & - & - & - \\
				& CharacterOnly & - & 7.05  & - & 95.08 \\
				\hline
				Hierarchical-stage & Pinyin-Character-Direct & 3.32 & 7.43  & 98.27 & 95.14  \\
				\hline
				Multi-task & Pinyin-Character-Multitask & 3.20 & 7.30  & 98.47 & 95.25  \\
				\hline
				Triangle & Pinyin-Character-Triangle & 2.75 & 6.90  & 98.39 & 95.98 \\
				\hline
				\multirow{2}*{Multi-decoder} & Pinyin-Character-Sequence & - & 6.91  & - & 95.47 \\
				& Pinyin-Character-Parallel  & 2.73 & 6.87  & 97.53 & 95.54 \\
				\hline
				\multirow{3}*{Interactive} & UnilateralAttention-Ch2Py & 2.76 & 6.70 & 97.80 & 95.61 \\
				& UnilateralAttention-Py2Ch & 2.81 & 6.83  & 98.00 & 95.66 \\
				& BilateralAttention & 2.68 & 6.83  & 98.52 & 95.78 \\
				\hline
				Proposed & Enhanced BilateralAttention & \textbf{2.65} & \textbf{6.60} & \textbf{98.55} & \textbf{95.98} \\
				\hline
			\end{tabular}
			\caption{CER on AISHELL-1 test sets compared to baseline and other variant models. \textbf{Pinyin Pred} means Pinyin transcript from prediction which measures pronunciation consistency. \textbf{Pinyin GT} means Pinyin transcript from ground-truth which measures pronunciation accuracy. \textbf{Pinyin-Character-Sequence}: Use the trained PinyinOnly model to initialize the CharacterOnly model and Retrain.}
			\label{proposedmodelresult_new}
			\vspace{-1.3cm}
		}
	\end{center}
\end{table*}

\subsection{Two-stage Training Strategy}

We train the Pinyin-Character dual-decoder model from a fix initialization to minimize the weighted sum of two cross-entropy loss:

\begin{equation}
\begin{aligned}
L_{combine}(\theta_{s}, \theta_{p}, \theta_{c}) = & \lambda L_{pinyin}(\theta_{s}, \theta_{p})  
\\
& + (1 - \lambda) L_{char}(\theta_{s}, \theta_{c})
\end{aligned}
\end{equation}
where $L_{pinyin}$, $L_{char}$ donate the cross-entropy loss of the pinyin decoder and character decoder, respectively. $\theta_{s}$, $\theta_{p}$, $\theta_{c}$, represent the parameters of the shared speech encoder, Pinyin decoder and Character decoder, respectively. $\lambda$ is a hyperparameter.  Here we adopt a two-stage training method:

\begin{algorithm}
	\renewcommand{\algorithmicensure}{\textbf{Output:}}
	\caption{Two-Stage Training}
	\label{alg1}
	\begin{algorithmic}[1]
	    \REQUIRE Speech $\textbf{s}$, Pinyin $\textbf{y}^{pinyin}$ and Character $\textbf{y}^{char}$
		\STATE Train Pinyin single model with $\textbf{s}$ and $\textbf{y}^{pinyin}$
		\STATE Train Character single model with $\textbf{s}$ and $\textbf{y}^{char}$
		\STATE Initialize dual-decoder model
		\STATE \quad Initialize encoder and Pinyin decoder with Pinyin single model
		\STATE \quad Initialize Character decoder with Character single model
		\STATE \quad Randomly initialize interaction module between decoders
		\STATE Re-train dual-decoder model with $\textbf{s}$, $\textbf{y}^{pinyin}$ and $\textbf{y}^{char}$
	\end{algorithmic}  
\end{algorithm}

\section{Experimental Details}

\subsection{Data and Evaluation Metrics}

In this paper, we use the AISHELL-1 dataset in 16 kHz WAV format to verify the performance of all the models. Each speech is represented as an 80-dimensional filterbank coefficients computed every 10ms with a 25ms window length. All feature sequences are normalized using the mean and variance of each audio sample. SpecAugment \cite{park2019specaugment} is employed to augment audio data for model training.

The Character-to-Pinyin conversion is done through the python-pinyin\footnote{https://github.com/mozillazg/python-pinyin} Python package. A Pinyin syllable consists of three parts: initial, final, and tone marker. In this paper, Pinyin syllables vocabulary size is 384 regardless of tone while the Character set consists of 4234 symbols including $<sos>$ $<eos>$ $<pad>$ tags.

In addition to using Character Error Rate (CER), we also design an evaluation metric to compare the alignment degree (AD) of Pinyin transcript and generated Character transcript for dual-decoder models which is defined as follows:
\begin{equation}
AD(\textbf{\textit{y}}_{char}, \textbf{\textit{y}}_{pinyin}) = \frac{\sum_{i}Same(\textit{y}^{convert}_{i,pinyin}, \textit{y}_{i,pinyin})}{len(\textbf{\textit{y}}^{convert}_{char}))}
\end{equation}
\begin{equation}
\textbf{\textit{y}}^{convert}_{pinyin} = F_{char \rightarrow pinyin}(\textbf{\textit{y}}_{char})
\end{equation}
\begin{equation}
Same(a, b) = \left\{ \begin{array}{rcl}
    1 & a = b \\
    0 & a \neq b
    \end{array} \right.
\end{equation}
where $F_{char \rightarrow pinyin}$ converts generated Character sequence to Pinyin sequence by mentioned automatic Python tool. The higher AD, the higher the correlation between Pinyin sequence and generated Character sequence.

\subsection{Model Details and Inference}
We adopt Conformer-Transformer \cite{gulati2020conformer,vyas2021optimally} which consists of Conformer Encoder \cite{vyas2021optimally} and vanilla Transformer Decode \cite{vaswani2017attention} as the basic architecture.
The baseline models include 6 encoder layers and 6 decoder layers. For a fair comparison, the SIDO dual-decoder model consists of 6 encoder layers, 3 decoder layers for Pinyin generation and another 3 decoder layers for Character generation, thus keep the parameter scale basically unchanged with baselines. The Conformer-Transformer consists of dimensional  $dim_{inner}$ = 2048, $dim_{model}$ = 512, $dim_{emb}$ = 512, $dim_{key}$ = 64 and $dim_{value}$ = 64. We use the Adam optimizer with a learning rate 0.001. Use the same optimizer settings as \cite{vaswani2017attention}. To further avoid overfitting, lable smoothing and dropout is used with probability 0.1. All models are trained for 80 epochs using a batch size 16. At test time, we use beam search with beam width 5 to output the best-decoded sentence.

\section{Experimental Results and  Analysis}

\subsection{Main Results}

The CER and AD scores of baselines and all the other related models on AISHELL-1 dataset are summarized in Table \ref{proposedmodelresult_new}. We can observe that the $\textit{Enhanced BilateralAttention}$ model with asynchronous dual-decoding, fuzzy Pinyin sampling and two-stage training achieves the best performance compared to all the other models. 
By comparing CER scores of different dual-decoder models, the results suggest that the dual-decoder models with layer-wise interaction is stronger than that with hierarchical interaction, i.e. $\textit{BilateralAttention}$ v.s. $\textit{Pinyin-Character-Triangle}$. Surprisingly, the performance of dual-decoder models sharing the whole or part of Pinyin decoder, i.e. $\textit{Pinyin-Character-Direct}$ and $\textit{Pinyin-Character-Multitask}$ is worse than the baseline models over Pinyin and Character CER which needs to be further explored.

By comparing AD scores of different dual-decoder models over Pinyin prediction, we can observe that the dual-decoder models with interaction is stronger than that without interaction, i.e. $\textit{BilateralAttention}$ v.s. $\textit{Pinyin-Character-Parallel}$.

Besides, we also draw the following empirical conclusions about Pinyin and Character for Mandarin ASR task.

\noindent \textbf{Pinyin ASR is much easier than Character ASR task.}
The PinyinOnly model achieves considerably lower CER scores compared to the CharacterOnly model, which is consistent with the analysis of Pinyin and Character for Mandarin Chinese ASR in Introduction section where Characters are logograms developed for the writing of Chinese while Pinyin is a coding system that spells out the sounds.

\noindent \textbf{Pinyin is a effective supplement.}
We find most of the dual-decoder models achieve competitive results, consistently surpass single decoder models of $\textit{PinyinOnly}$ and $\textit{CharacterOnly}$. 

\noindent \textbf{Pinyin ASR is an effective pre-training task for Character ASR task.}
$\textit{Pinyin-Character-Sequence}$ achieves better results compared to the CharacterOnly model trained from scratch. It shows that Pinyin pre-training provides a better model initialization.

\subsection{Effect of initialization strategy}

To explore how different initialization strategies contribute to performance, we conduct an ablation experiment with $\textit{UnilateralAttention}$ and $\textit{BilateralAttention}$ model. As is shown in Table \ref{initialization}, we compare with several different initialization approaches. We can find that all initialization methods with single Pinyin/Character model is only benefit to Pinyin/Character side of dual-decoder model and two-stage training is more critical. 

\begin{table}[ht]
	\begin{center}
		\begin{tabular}{c | c | c | c | c}
			\hline
			\multirow{2}*{Initialization} & \multicolumn{2}{c |}{Unilateral-Py2Ch}  & \multicolumn{2}{c}{Bilateral} \\
			\cline{2-5}
			& Pinyin & Character & Pinyin & Character \\
			\hline
			Random & 2.81 & 6.83 & 2.68 & 6.83 \\
			Single Pinyin & 2.59 & 6.98 & 2.60 & 6.88 \\
			Single Character & 2.82 & 6.80 & 2.81 & 6.58 \\
			Two-Stage & 2.71 & 6.73 & 2.63 & 6.62 \\
			\hline
		\end{tabular}
		\caption{Comparison of different training strategy. \textbf{Random}: training from scratch. \textbf{Single Pinyin}: Initialize encoder and Pinyin decoder with pretrained PinyinOnly. \textbf{Single Character}: Initializa Character decoder with pretrained CharacterOnly}
		\vspace{-1.0cm}
		\label{initialization}
	\end{center}
\end{table}

The advantage of the two-stage training method is that the dual-decoder model can get a better model initialization from separate models while speeding up convergence and model training.


\subsection{Effect of fuzzy Pinyin sampling probability}
To investigate the benefits of fuzzy Pinyin training, we compare the CER results of different sampling probability $\textit{p}$ with $\textit{BilateralAttention}$ model. As is shown in Table \ref{Pinyin_sampling_probability}, we gradually increase the sampling probability $\textit{p}$ to assess the impact. There are increases in the performance improvement as the increase of sampling probability increases. However, performance will degrade significantly when exceeding the threshold. We suspect that this is due to the introduction of excessive misinformation and increased inconsistency between training and testing.

\begin{table}[ht]
	\begin{center}
		\setlength{\tabcolsep}{8.0mm}{
		\begin{tabular}{c | c |  c}
			\hline
			\makecell[c]{\multirow{2}*{\textit{p} (\%)}} & \multicolumn{2}{c}{BilateralAttention CER(\%)} \\
			\cline{2-3}
			& Pinyin & Character  \\
			\hline
			\makecell[c]{0} & 2.77 & 6.83 \\
			15 & 2.74 & 6.78 \\
			20 & 2.69 & 6.65 \\
			30 & 2.87 & 6.97 \\
			\hline
		\end{tabular}
		}
		\caption{Fuzzy Pinyin sampling probability p study}
		\vspace{-1.0cm}
		\label{Pinyin_sampling_probability}
	\end{center}
\end{table}

It is worth noting that the optimal probability value is also depending on the choose of model type, training data, similar Pinyin confusion set and so on.

\subsection{Effect of hyperparameter $\lambda$}

To analysis how weight $\lambda$ of the combined loss contributes to performance with $\textit{BilateralAttention}$ model, we conduct an experiment. 

\begin{table}[ht]
	\begin{center}
		\setlength{\tabcolsep}{8.0mm}{
		\begin{tabular}{c | c | c}
			\hline
			\multirow{2}*{$\lambda$} & \multicolumn{2}{c}{BilateralAttention CER(\%)} \\
			\cline{2-3}
			& Pinyin & Character \\
			\hline
			0.2 & 2.94 & 6.84 \\
			0.5 & 2.68 & 6.83 \\
			0.8 & 2.90 & 7.39 \\
			\hline
		\end{tabular}
		}
		\caption{Effect on performance of loss weight $\lambda$}
		\label{hyperparameterresult}
		\vspace{-1.0cm}
	\end{center}
\end{table}

As is shown in Table \ref{hyperparameterresult}, we find that the higher weight on the Character side is more helpful. Because generating Character transcripts is more difficult than generating Pinyin transcripts, it is reasonable for the character side to have a higher weight. However, ignoring the Pinyin side will harm the overall performance of the model. In this paper, we set $\lambda$ to be 0.5.




\section{Conclusion}

In the paper, we introduce the variants of Pinyin-aware Mandarin Chinese ASR architectures in SIDO frameworks and propose an asynchronous dual-decoding with fuzzy Pinyin sampling method for Mandarin Chinese ASR task utilizing the one-to-one corresponding characteristics between Pinyin and Character. What's more, we design an effective two-stage training strategy to get a better initialization. The results on the test set of AISHELL-1 dataset show that the proposed model outperforms various related models and mainstream ASR models on AISHELL-1 dataset. 

\bibliographystyle{IEEEtran}

\bibliography{template}
\end{document}